\documentclass{article}

\usepackage{neurips_2025}

\usepackage[utf8]{inputenc} 
\usepackage[T1]{fontenc}    
\usepackage{hyperref}       
\usepackage{url}            
\usepackage{booktabs}       
\usepackage{amsfonts}       
\usepackage{nicefrac}       
\usepackage{microtype}      
\usepackage{xcolor}         
\usepackage{graphicx}
\usepackage{multirow}
\usepackage{amsmath}
\usepackage{comment}

\title{Fast Fourier Correlation is a Highly Efficient and Accurate Feature Attribution Algorithm from the Perspective of Control Theory and Game Theory}

\author{%
  Zechen Liu\\
  School of Computer Science\\
  Wuhan University\\
  \texttt{zecliu@whu.edu.cn} \\
  \And
   Feiyang Zhang \\
  Brain Research Center \\
  Second Clinical School\\
  Wuhan University \\
  \texttt{fyzhang.neu@whu.edu.cn} \\
  \AND
  Wei Song \\
  School of Computer Science \\
  Wuhan University \\
  \texttt{songwei@whu.edu.cn} \\
  \And
  Xiang Li \\
  Brain Research Center \\
  Second Clinical School\\
  Wuhan University \\
  \texttt{li.xiang@whu.edu.cn} \\
  \And
  Wei Wei \\
  Brain Research Center \\
  Second Clinical School\\
  Wuhan University \\
  \texttt{wei.wei@whu.edu.cn} \\
}

\begin{document}

\maketitle

\begin{abstract}
Understanding the decision-making process of neural networks remains elusive, with attribution analysis being a key tool. In the spatial domain, despite significant progress, attribution methods still lack a principled and widely accepted correctness evaluation metric, making it difficult to rigorously evaluate their reliability. Meanwhile, in the Fourier domain, recent works have revealed that neural networks exhibit a strong bias toward features with low frequency, but have yet to yield an attribution method that is capable of isolating individual features with the same granularity as in the spatial domain.
In this paper, we first leverage the definition of deletion in the field of signal filtering to justify the use of game theory to find an evaluation metric in the Fourier domain. Building on this, we introduce a fast Fourier correlation(FFC) attribution algorithm inspired by closed-loop feedback control systems.
Empirical results demonstrate that FFC significantly outperforms existing spatial attribution techniques in the game theory-based metric. With the ImageNet-2012 dataset, we show that just $4\%(6,021/150,528)$ of the Fourier features identified by FFC are sufficient to preserve the original predictions of ViT-B/32 for $80\%$ of samples. Furthermore, we find that high-score Fourier features found by FFC exhibit strong intra-class concentration and stable inter-class specificity. These findings highlight the potential of Fourier-domain attribution as a powerful tool for interpretable and robust AI.

\end{abstract}

\section{Introduction}
Attribution analysis has emerged as one of the primary tools for interpreting the decision basis of neural networks. Among various approaches, spatial attribution methods have gained widespread popularity due to their ability to produce heatmaps that align with human visual intuition (\cite{gradcam}). Despite extensive studies, spatial attribution still lacks a universally accepted evaluation metric for correctness (\cite{LPI, Quanshi}). Notably, different attribution methods based on spatial domain often perform inconsistently(\cite{aggattri}), and even produce contradictory explanations for the same model and input(\cite{conflict1,conflict2,unstabletheo22,unstablenips2020,unstabletheo23,unstablenips}).

Most of the evaluation metrics (\cite{metricassumption,metricassumtionnips})are built upon a consensus: perturbing important features should significantly alter the model's output, whereas perturbing unimportant ones should have minimal effect (\cite{fullgrad}). However, using such a principled perturbation operation in the spatial domain to independently perturb a signal remains an open question. A commonly used strategy is to remove/delete the selected pixel (set the value of selected pixels to zero or a certain value(\cite{LPI,DiffID2,fullgrad,IR})), but the effect of the removed pixels still exists. It often introduces high-frequency artifacts that obscure the original intent of removing/deletion(\cite{attbenchmark}). Some efforts have attempted to formalize perturbations through axiomatic design(\cite{intgrad,infd}), yet no universally accepted axioms have emerged. In contrast, in the Fourier domain, it is rather easy to independently perturb (or remove) a signal because the Fourier transform provides a mathematically and physically grounded way to represent a sample as a superposition of frequency components. Specifically, the commonly used remove/delete strategy can be defined by zeroing the energy of a component, which is a well-established deletion operation according to the relevant signal filtering process(\cite{ampli-filter,Conv}), thereby eliminating the artifact-related ambiguities that plague spatial attribution. Moreover, game theory-based metrics such as the Deletion-Insertion Game can be more reliably applied in the Fourier domain. 

In recent years, we have witnessed a surge in Fourier-domain approaches for analyzing neural networks(\cite{zhiqinxuintroDFprinciple,frequencydomaintrainingbehavior,vitlowfre,vitgood,Fprinciple,icmlciteFpreciple,frequency1robustness}), many of which have influenced downstream applications like model compression(\cite{modelcomression}). A growing body of work has shown that neural networks tend to prioritize low-frequency features(\cite{shapFre,vitlowfre,NTKFourier}). However, there remains a lack of attribution methods that can identify individual Fourier features relevant to model predictions, especially for modern network architectures.

Motivated by the limitations of spatial attribution and the absence of fine-grained Fourier attribution, we propose an efficient Fourier Feature Attribution method grounded in control theory called \emph{Fast Fourier Correlation (FFC)} attribution algorithm, which ranks and scores the frequency components based on the extend of contributions for the decisions of a neural network.

Our experiments show that removing the low-score features identified by FFC \textbf{increases} the model’s confidence in the original class, indicating \textbf{noise removal}. Conversely, removing high-score features leads to a rapid drop in corresponding confidence, confirming their relevance. Compared to spatial attribution, our approach removes noise more effectively while preserving less redundant information. Additionally, we find that high-score Fourier features exhibit significantly higher intra-class concentration and stable inter-class specificity, suggesting that Fourier-based attribution offers a promising pathway toward more interpretable and structured classification AI systems.

\textbf{Main Contributions:}
\begin{itemize}
    \item 1. We \textbf{formalize the deletion operation in the Fourier domain} using concepts from signal filtering and game theory. We demonstrate, both theoretically and physically, why game theory-based evaluation metrics behave differently in the spatial and Fourier domains and why the Fourier domain allows for a mathematically well-defined and artifact-free deletion process.
    \item 2. We \textbf{propose a novel Fourier feature attribution algorithm} based on feedback control theory in closed-loop systems. Leveraging the correlation properties among frequency components, we formulate a projection-based attribution scheme. Experiments show our method achieves stronger feature selectivity compared to existing attribution approaches.
    \item 3. We \textbf{find characteristics of high-score Fourier features}. Compared to existing attribution approaches, our results reveal that high-score Fourier features are fewer in number but demonstrate superior intra-class concentration and comparable inter-class specificity. This indicates that Fourier-based attribution may serve as a more structured and scalable foundation for interpretable classification AI.
\end{itemize}
\section{Related Works}
\subsection{Studies of Spatial Attribution Methods and Metrics}
\textbf{Attribution methods}: Spatial attribution methods have achieved significant attention for their ability to generate intuitive heatmaps that align with human perception (\cite{inputgrad}. A wide range of techniques has been proposed based on different theoretical underpinnings. These include gradient-based approaches: Integrated Gradients (\cite{intgrad}, LPI (\cite{LPI}, MIG (\cite{MIG}, and Expected Gradients (EG) (\cite{EG}; perturbation-based: SmoothGrad (\cite{smoothgrad} and LIME (\cite{LIME}; backpropagation-based: DeepLIFT (\cite{deeplift}, Input×Gradient (\cite{inputgrad}, Guided Backpropagation (\cite{gbp}, FullGrad (\cite{fullgrad}, and Grad-CAM (\cite{gradcam}. More recently, hybrid methods that integrate multiple attribution paradigms have also been proposed (\cite{aggattri}.\\
\textbf{Metrics}:
Despite their diversity and intuitive appeal, spatial attribution methods still lack a universally accepted metric for evaluating correctness (\cite{Quanshi,fullgrad,LPI,infd,withBenchmark1}). Moreover, the existence of counterfactual samples (\cite{harvardTrajectory}) raises concerns about the reliability and stability of spatial attribution explanations (\cite{unstable,unstablenips,unstableAAAI,unstabletheo22,unstabletheo23,unstablenips2020}. For instance, it has been shown that some attribution methods yield contradictory explanations for the same model and input (\cite{conflict1,conflict2}, and similar doubts have been raised regarding their consistency across perturbations (\cite{unstable,unstabletheo23}.
To address these issues, numerous evaluation metrics have been proposed to assess the faithfulness of spatial attribution. These include FID(\cite{FID}), IR(\cite{IR}), INFD (\cite{infd}), ROAR(\cite{attbenchmark}), DIFFID(\cite{DiffID2}), and the Deletion-Insertion Game(\cite{fullgrad}). Broadly, these metrics fall into two categories: game theory-based approaches and axiom-based approaches. While the former provides a principled framework for evaluating feature importance, it suffers from artifacts introduced by deletion operations in the spatial domain(\cite{LPI,fullgrad,craftartifact,TCAV,FromTCAV}). The latter, on the other hand, often lacks consensus on which axioms should be considered foundational, resulting in fragmented and sometimes incompatible criteria.
To overcome the limitations of spatial attribution, we suggest shifting the attribution process into the Fourier domain. Because the Fourier transform decomposes an input into a linear combination of orthogonal frequency components. This decomposition enables a mathematically and physically grounded definition of deletion operations, i.e., zeroing out signal energy, a well-established procedure in classical signal processing (\cite{ampli-filter,Conv}). As a result, game theory-based evaluation metrics like the Deletion-Insertion Game can be implemented in the Fourier domain without introducing artifacts. This insight motivates our use of Fourier-based attribution as a more principled and robust alternative to spatial methods.
\subsection{Studies of Analyzing the networks via Fourier Features}
A growing body of works has explored neural networks through the lens of the Fourier domain, with the central finding being that modern networks tend to capture low-frequency components(\cite{vitlowfre,wang2022antioversmoothing}). Numerous studies(\cite{vitgood,vitlowfre,NTKFourier,shapFre}) have pointed out that low-frequency features may generalize better, and results from compression theory(\cite{compresstheory}) and model compression(\cite{modelcomression}) further support the idea that low-frequency components carry more information than their high-frequency counterparts.
In addition, recent work has demonstrated that convolutional layers exhibit implicit frequency-selective behavior (\cite{wen2024which}. However, this theoretical result is derived under idealized assumptions, such as convolution kernels being the same size as the input conditions and the network is performing recurrent convolution, that rarely met by contemporary architectures like Vision Transformers (ViTs) and ResNet.
While these studies have laid important groundwork, they fall short of offering practical tools for attributing individual Fourier features to model decisions in modern networks. In light of this gap, we propose a FFC that generalizes to contemporary architectures. Our approach builds on prior insights and extends the analytical power of Fourier-based interpretations into a practical and scalable attribution framework.

\section{Method}
\subsection{Deletion Insertion Game}
Due to the established game theory, we select the deletion-insertion game to be our backbone metric. However, in the spatial domain, game theory-based metrics are criticized because they often involve new artifacts, making it hard to independently remove a signal, which violates the basic assumption of game theory. After the Fourier transformation, the sample is represented by a linear combination of signals. Thus, setting the corresponding energy to zero aligns with the definition of deletion in signal filtering without influencing other orthogonal components(\cite{ampli-filter}). For a sample with shape $m\times n$, the aforementioned procedure could be described as below:
\begin{align}
    F_X\left(u,v\right) &= \sum\limits_{x}^{m}\sum\limits_{y}^n X\left(x,y\right)e^{-j2\pi\left(ux+vy\right)}\\
    F_{X} &= \sum\limits_u^m\sum\limits_v^n F_X\left(u,v\right)\\
    F_{X}^{del} & = F_X-F_X\left(u^{del},v^{del}\right),
\end{align} where $F_X$ denotes the signal in the Fourier domain with the direction of $\left(u,v\right)$ with a frequency magnitude $\sqrt{u^2+v^2}$. $X$ denotes the sample in the spatial domain. $F_X$ denotes the sample in the Fourier domain and $F_{X}^{del}$ denotes the sample after deleting the particular signal.
From the viewpoint of mathematics, zeroing a frequency component removes a Fourier item. From the viewpoint of signal filtering, the influence of the deleted frequency component is removed. However, in the spatial domain, setting the value of a pixel to a certain value is adding a pulse signal from the perspective of physics, thus involving a series of new Fourier items from the perspective of mathematics. The influence of the deleted pixel still exists.

Grounded in the theoretical foundations of the definition of the removal in signal filtering(\cite{Conv,ampli-filter}), we argue that applying the game theory-based Deletion-Insertion Game in the Fourier domain is both mathematically and physically well-justified.
\begin{figure}
    \centering
    \includegraphics[width=0.9\linewidth]{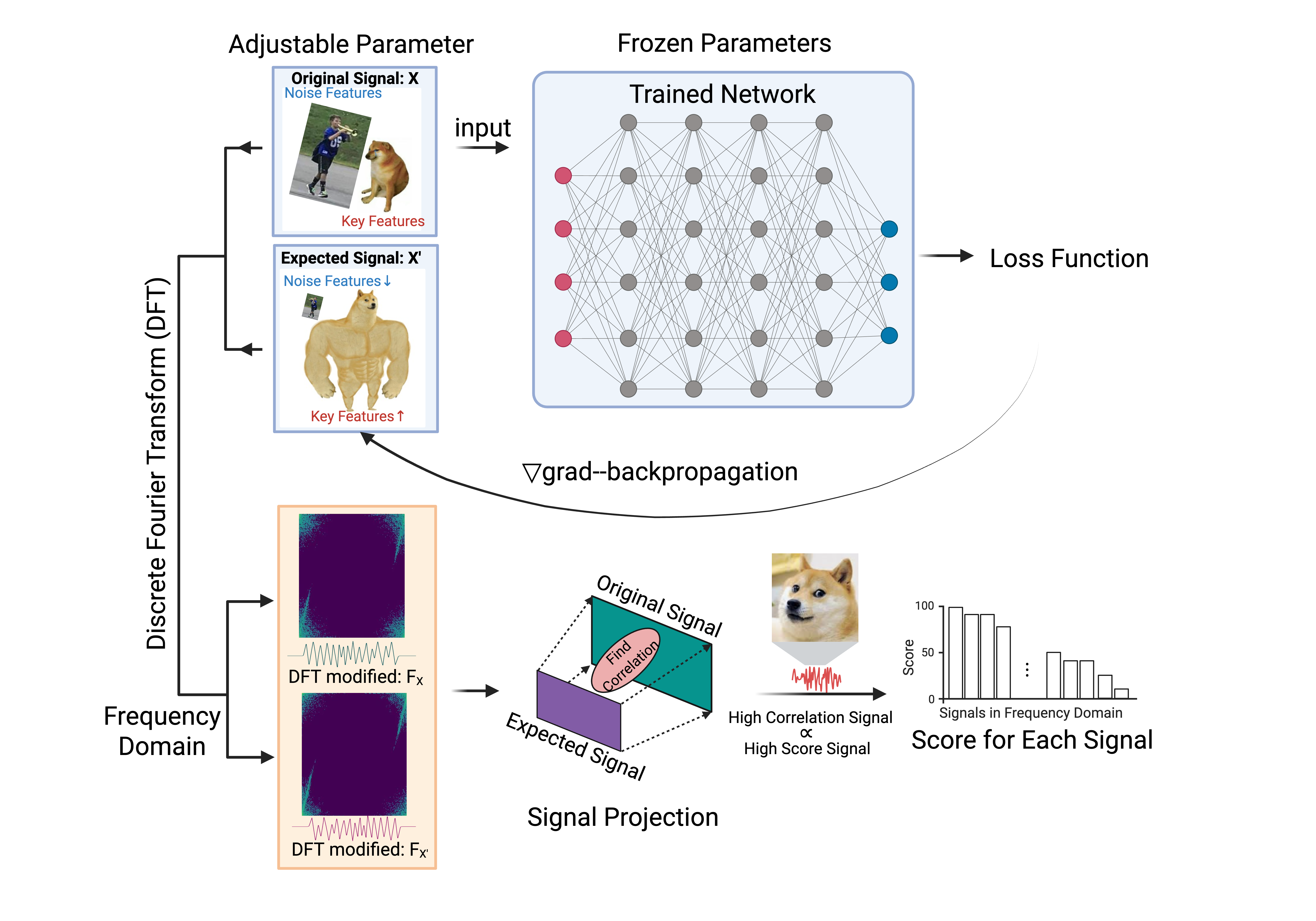}
    \caption{\textbf{The Schematic workflow of FFC}: Original Signal $X$ is regarded as an adjustable parameter and input to a trained neural network with all parameters being frozen. Loss function is leveraged to revise the Original Signal $X$ according to the preferences of the neural network, yielding the Expected Signal $X'$. The pair of signals will then go under a Discrete Fourier Transform and be changed from the spatial domain to the Fourier domain, denoted as $F_{X}$ and $F_{X'}$, respectively. Further in the Fourier domain, the expected signal will be projected to the original signal, and a measurement of correlation will be conducted. Finally, according to the correlation of each signal pair, we can score and rank the signals in the Fourier domain.}
    \label{fig:method}
\end{figure}
\subsection{Fast Fourier Correlation Algorithm}
Building upon our Fourier-domain adaptation of the evaluation metric, we further model the attribution process as a control process of the closed-loop feedback system. While the idea of modeling networks using control theory dates back to early works on Hopfield networks(\cite{neurondynamics}), to the best of our knowledge, control-theoretic formulations have not yet been applied to feature attribution.

Unlike in network training, where the control parameters are typically the network weights(\cite{analyneurondynamics1}), the attribution process respectively treats the input sample and the model’s output as controllable and observable states. As illustrated in Figure~\ref{fig:method}, we freeze the weights of the target model and treat the attribution process as a dynamic system that iteratively adjusts the input to minimize the discrepancy between the network’s current output and a target output.

As shown in Figure~\ref{fig:method}, we regard the cross-entropy loss function as a controller, using the loss value as the control signal. Gradients computed from the network’s output are propagated backward to the input layer, acting as error signals, which guide subsequent updates. We implement this mechanism via gradient descent, enabling real-time estimation of feature importance in the Fourier domain.
Then we adopt the gradient descent algorithm and error signals to rectify the input signals and eliminate the loss value. We refer to the modified input signals that drive the network output into a steady state (minimizing the loss value) as the "expected signal". According to the control theory, we assume that the original signals contain expected signals and noise signals, and the expected signals are enhanced during the modification process while the noise signals are reduced or modified greatly. Therefore, the variance between Original Signals and Expected Signals reflects the importance of signals. To independently evaluate the variance of each signal, we perform the Fourier transformation on the pair of samples and represent them in K-space. We calculate the projection of each component in the expected signals onto the corresponding components in the original signals. The projection formula is given below:
\begin{align}
    F_{X'} &=\mathcal{F}\left(X^{\left(n\right)}-lr\cdot\nabla X^{\left(n\right)}\right)\\
    Proj\left(u,v\right) &= \frac{Re\left(F_X\left(u,v\right)\cdot\overline{F_{X'}}\left(u,v\right)\right)}{En\left(F_{X'}\left(u,v\right)\right)},
\end{align} where $lr$ denotes the learning rate, $X^{\left(n\right)}$ denotes the rectified input after $n$ iterations of gradient descent algorithm, $X^{\left(n\right)} - \nabla X^{\left(n\right)}$ denotes the Expected Signals $X'$, $\mathcal{F}$ means Fourier transformation, $\overline{F_{X'}}$ means the conjugation signal of the expected signal $F_{X'}$ and the $Re\left(\cdot\right)$ means the real part of the complex number result, $En\left(\cdot\right)$ denotes the magnitude of the signals. And then we use the value of the projection minus the magnitude of the original signal to obtain the importance score:
\begin{equation}
    Importance\left(u,v\right) = Proj\left(u,v\right)-En\left(F_X\left(u,v\right)\right),
\end{equation}

\section{Experiment}
Our experiments are designed to answer the questions: \textbf{1)}Are the expected signals enhanced while noise signals are weakened? How do parameters such as learning rate and iteration influence the results? \textbf{2)} What does the performance of FFC differ from traditional spatial domain attribution methods using Deletion Insertion Game as evaluation? \textbf{3)} How does the computational overhead of FFC compare with existing attribution approaches? \textbf{4)} What are the characteristics of high-score Fourier features? Are they suitable for interpretable classification AI?
\begin{figure}
    \centering
    \includegraphics[width=1\linewidth]{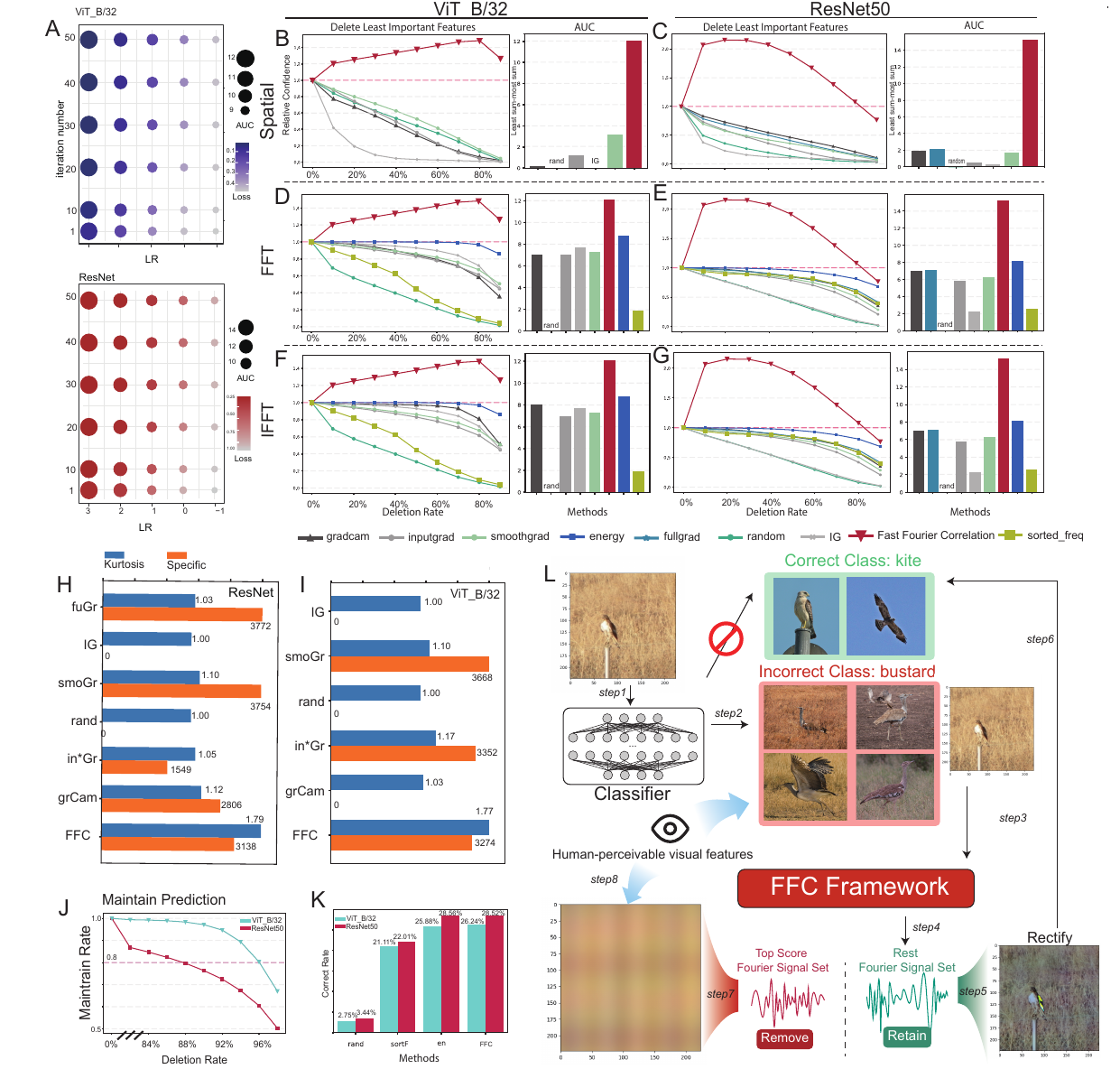}
    \caption{Figures for Experiments.(A) Parameter Analysis. The x-axis LR represents log learning rate, values ranging from $0.1$-$1000$(refer to section~\ref{Param-Ana} for reasons.). The y-axis means iteration times, ranging from $1$-$50$. The values of Loss and AUC are represented by the color and size of the dots. The parameter analysis reveals that performance is primarily influenced by the learning rate. (B-C) Direct comparison experiments of different attribution methods. Left panel: the x-axis represents the percent of low-score features being filtered out, the y-axis represents the confidence of the network. Right panel: the x-axis indicates attribution methods, the y-axis represents the AUC. (D-G) Comparison experiments in the Fourier domain. Axis meanings are the same as B-C. The study in the Fourier domain suggests FFC’s superiority is not similar to any existing method and surpasses the simple FFT/IFFT transform of existing attribution methods' results. (H, I, J) Characteristic analyses. H, I: shows the kurtosis and the specificity of the high-score Fourier Features and spatial features, revealing that the high-score Fourier features hold higher intra-class concentration and stable inter-class specificity. J: shows the maintain rate of $2$ models when deleting features ranked by FFC. (K) Success rate of the rectification experiment. (L) Schematic pipeline of the procedure for understanding the visual manifestation of Fourier features through misclassified sample analysis. Step 1: input sample to the network; step 2: the network misclassifies the sample into category of bustard; step 3: rank the importance of the features using FFC; step 4: remove the top $1/10,000$ features; step 5: visualize the rectified sample; step 6: the network correctly classifies the sample into the right category kite; step 7: visualize the deleted features; step 8: find the correlation between the deleted features and the category of bustard. The deleted features are more similar to the yellow background that appears in the category of bustard.}
    \label{fig:experiment}
\end{figure}
\subsection{Settings}
\textbf{Dataset}: Validation set of the Imgnet2012(\cite{imgnet2012}) dataset with $1000$ categories($50$ samples per category). It is used in many previous works(\cite{fullgrad,craftartifact,LPI,gradcam}). The complete Imgnet2012 validation set is interpreted in terms of individual baselines.\\
\textbf{Baselines}: Integrated-based: IntGrad(\cite{intgrad}), Perturbation-based: SmoothGrad(\cite{smoothgrad}), Back-Propagation: GradCAM(\cite{gradcam}), FullGrad(\cite{fullgrad}). We select baselines from different types of foundations.\\
\textbf{Implementation details}: Experiments are conducted with A800 80GB and CentOS 8. The IntGrad is implemented by captum (\cite{captum}). The baseline value is set to zero. Other baselines are implemented by open-source code of (\cite{fullgrad}). All parameters are set according to the default parameters from open-source code and the relevant research paper. The discrete Fast Fourier transforms (FFT) are implemented by PyTorch's native torch.fft package. All experiments were repeated three times to take the final average. \textbf{NOTE: Except for Smoothgrad, the other methods DO NOT contain random variables, thus the errors are too small. We use the best result of the Smoothgrad to conduct experiments}. For details on the error, refer to our supplementary. As for Fullgrad, because we didn't retrain the ResNet, the accuracy of the official ResNet-50 is around $80\%$, the result is different from their original paper(their accuracy is above $95\%$). \textbf{Fullgrad does not support ViT\_B/32.}\\
\textbf{The backbone networks}: ResNet-50 and ViT-B/32, both use the official PyTorch pre-trained models with default initialization weights.
For more details, please refer to the Supplementary
\subsection{Parameter Analysis}
\label{Param-Ana}
FFC is influenced by the learning rate and the number of iterations. This section analyzes the impact of the two parameters on the expected output of the loss value and the final Deletion Insertion Game. The relative confidence is computed by modified confidence divides the original confidence. Because the norm of the gradient is ignored, our $lr$ ranges from $0.1$ to $1000$(\cite{GD}). The iteration ranges from $1$ to $50$. As shown in Figure~\ref{fig:experiment}.A, instead of producing instability, the loss value decreases quite rapidly when LR increases. Moreover, with increasing iteration number, loss of all learning rates shows a stable decreasing trend as well, even with $lr>1$. This illustrates the stability of FFC. 
As shown in Figure~\ref{fig:experiment}.A, the AUC value of the Deletion Insertion Game curves gets larger as the loss decreases (The AUC value is defined as area under the curves of \emph{deleting the least important features} minus the area under the curves of \emph{deleting the most important features}). The curves of deleting the most important features are contained in the supplementary. This suggests that the expected signals are amplified and the noise signals are weakened or modified by FFC, corresponding with our previous hypothesis. We adopt $lr=1000, e=50$ to conduct the rest of the experiments.
\subsection{Direct Comparison Experiments}
To compare the differences between FFC and previous spatial attribution algorithms under the Deletion Insertion Game metric in the spatial domain directly, we sequentially set the pixel value in the spatial domain to zero(the widely used remove operation) according to the scores from the spatial attribution algorithms. To ensure the removing consistency between $2$ domains, the numbers of parameters (pixels in spatial domain/signals in Fourier domain) set to zero are equal. Thus, for FFC, we set the same amount of signal energy to zero according to the scores as in the spatial domain.
As shown in Figure~\ref{fig:experiment}.B-C, removing the low-score features identified by FFC can increase the network's confidence. This indicates that the low-score features identified by FFC are noise that confuses the network's decision-making. In contrast, none of the spatial attribution algorithms can achieve an increase in network confidence. Moreover, even though FFC removes up to $90\%$ of the signals, the network's relative confidence is still around $1$. This demonstrates that FFC can eliminate most of the noise while retaining the important information. This fully illustrates that FFC significantly outperforms spatial attribution algorithms in feature selection capability. Additionally, the AUC value significantly surpasses the baselines which demonstrates that FFC is more faithful.
\subsection{Comparison Experiments in Fourier Domain}
This section further highlights the novelty of FFC. To demonstrate that FFC outperforms the results of directly applying fast Fourier transform (FFT) or inverse FFT (IFFT) to the baseline attribution methods, we transform the attribution result of the baseline methods to the Fourier domain and apply the Deletion Insertion Game to evaluate them. Baseline methods include algorithms that delete features based on frequency magnitude, random scores, signal magnitude,  FFT and IFFT of the scores of various spatial domain algorithms. As shown in Figure~\ref{fig:experiment}.D-G, the attribution results of FFC significantly outperform all the baseline methods. More specifically, to test the difference of FFC from the compression technique(\cite{compresstheory,modelcomression}) and the existing analysis(\cite{vitlowfre,Fprinciple}), we conduct the comparison to the sorted\_freq, surpassing indicates a better performance than the compression technique. Meanwhile, the surpassing compared to the energy baseline method demonstrates our difference from the previous work which claims energy bias causes the network preference(\cite{shapFre}). FFC's surpassing compared to the FFT/IFFT of the current spatial algorithm demonstrates that the results of FFC cannot be simply derived from existing results through trivial transformations. These results suggest that FFC's novelty compared to existing attribution methods.
\subsection{Computational Overhead}
To compare the computational overhead of FFC with existing attribution algorithms, we fix the batch size equals $128$ to compare the spatial and temporal overheads. For methods that exceed the device limits, we use the maximum batch size that we can handle. The spatial overhead of the algorithms is measured by the peak memory usage during runtime, while the temporal overhead is measured by the average time required to compute one batch. The representation method is average time × iteration(e) times. To eliminate errors caused by system I/O, the time cost is measured from the moment the data enters the GPU.
\begin{table}[]
    \centering
    \caption{Computational Overhead}
    \label{tab:cost}
    \begin{tabular}{c|c|c|c|c}
       Methods  & Batch Size & Time & Space(MB) & Model \\
       \hline
       Intgrad  & 16 & 1.422s/Batch*3125 Batches & 73010 & \multirow{8}*{ResNet50} \\
       Fullgrad  & 32 & 0.124s/Batch*1563 Batches & 51278 & ~ \\
       smoothgrad  & 128 & 8.521s/Batch*391 Batches & 12212 & ~ \\
       Inputgrad  & 128 & 0.067s/Batch*391 Batches & 12004 & ~ \\
       Gradcam  & 128 & 0.087s/Batch*391 Batches & 13122 & ~ \\
       FFC(e$=1$,AUC$=15.10$)  & 128 & 0.071s/Batch*391 Batches & 14064 & ~ \\
       FFC(e$=10$,AUC$=15.14$)  & 128 & 1.16s/Batch*391 Batches & 14064 & ~ \\
       FFC(e$=50$,AUC$=15.23$)  & 128 & 6.37s/Batch*391 Batches & 14064 & ~ \\
       \hline
       Intgrad  & 32 & 2.093s/Batch*1563 Batches & 53206 & \multirow{7}*{ViT\_B\_32} \\
       smoothgrad  & 128 & 16.135s/Batch*391 Batches & 5422 & ~ \\
       Inputgrad  & 128 & 0.067s/Batch*391 Batches & 4908 & ~ \\
       Gradcam  & 128 & 0.087s/Batch*391 Batches & 5684 & ~ \\
       FFC(e$=1$,AUC$=11.75$)  & 128 & 0.081s/Batch*391 Batches & 6490 & ~\\
       FFC(e$=10$,AUC$=11.84$)  & 128 & 2.038s/Batch*391 Batches & 6490 & ~\\
       FFC(e$=50$,AUC$=12.09$)  & 128 & 11.22s/Batch*391 Batches & 6490 & ~\\
    \end{tabular}
    
\end{table}
Compared to the baselines, the time cost of FFC is second only to input*grad when the number of iterations ($e$) is $1$. As displayed in Figure~\ref{fig:experiment}.A and Table~\ref{tab:cost}, the AUC value is similar to the $e=50$(changing rate less than $4\%$). The memory cost of FFC is also close to the baselines, except for Intgrad and fullgrad. Since the computational overhead of back-propagation is close to that required for model training, we argue that FFC has potential for industry applications.
\subsection{Fourier Feature Characteristic Analysis}
To study the characteristics of Fourier features found by FFC, this section analyzes the  FFC attribution results by comparing them to existing spatial results from the perspective of intra-class concentration and inter-class specificity. The high-score features of different algorithms are defined as features with scores exceeding the mean score within each sample. Further, to eliminate the difference of scales across methods, we tag the high-score feature to $1$ and the low-score to $0$. \\
\textbf{Intra-class Concentration}:
Kurtosis is used to measure the concentration, which reflects the ability of attribution methods to filter out noise. The higher the kurtosis, the stronger the noise-filtering capability. Within each class, feature tags are added up across all samples with the same label along each dimension to compute an overall score. We then calculate the kurtosis of the high-score features. The results in Figure~\ref{fig:experiment}.H-I indicate that Fourier features exhibit a higher concentration, with both kurtosis values approaching $1.8$ for RestNet and Vit-B/32. In contrast, spatial attribution methods exhibit lower concentration, barely exceeding random attribution by $0.2$ at best.
\textbf{Inter-class Specificity}: 
To assess inter-class specificity, the mean appearance times of high-score features across different classes are computed. Given the binarization of feature tags to $1$ for high and $0$ for low, a feature’s inter-class mean value approaching $1/1,000$ indicates greater specificity, given that $1/1000$ represents the feature is only presented in one class. We count the number of features with an inter-class mean equal to $1/1000$, as a higher count suggests higher specificity. Figure~\ref{fig:experiment}.H-I show that while FFC's kurtosis is significantly higher than any baselines, its inter-class specificity metric remains stable above $3000$ in both models, ranking in the top $3$ among all methods. This indicates that the features found by FFC are more suitable for classification which aligns with our claim.\\
\textbf{Minimum Number of Features to Maintain Decision}:
The comparison experiment from the perspective of confidence demonstrates that FFC can identify a large amount of noise in the samples. This experiment, from the perspective of accuracy, illustrates the minimum number of features required to maintain the original decision of the samples. We define maintaining the decision as the highest-confidence class of the processed sample being consistent with that of the original sample. As shown in Figure~\ref{fig:experiment}.J, for both backbone networks, only a small number of Fourier features are needed to maintain the original decision. For the ViT model, just $4\%$ of the Fourier features are sufficient to maintain the original decision for $80\%$ of the samples, while ResNet requires slightly more, at $15\%$ of the features to maintain the original decision for $80\%$ of the samples.
\subsection{Understanding the Visual Effects of Fourier Features}
To visually understand the visual effects of Fourier features attribution, we select Fourier features that cause misclassifications and transform them back to the spatial domain for interpretation. Specifically, using the FFC, we identify high-score features in misclassified samples(approximately $20\%$ of the samples) and sequentially remove the top-scoring features (we assume they are the most important features that cause misclassifications), ranging from $1/10,000$ to $1/1,000$, with a step size of $1/10,000$. The process stops when the sample's classification is corrected (i.e., the highest-confidence class matches the true label). A correction is considered failed if the sample can not be corrected after removing the top $1/1,000$ of the features. We systematically compare the correction rates of FFC to those of the baseline methods (random removal, removal based on frequency magnitude, and removal based on energy magnitude). Figure~\ref{fig:experiment}.I shows that the correction rate of FFC ranges from $25\%$ to $28\%$, which is significantly higher than the random removal($3\%$) and the frequency magnitude-based removal($20\%$-$22\%$). However, there is no significant difference compared to the energy-based removal method. Due to the fact of simply removing top features instead of finding all features causing the misclassification, FFC still corrects errors to a certain extent and holds considerable potential for future practical applications.

Figure~\ref{fig:experiment}.L illustrates an example from the dataset, the $42$-th image of class $21$ kite. This image is misclassified by the ResNet50 to the class of $138$, bustard. The image is selected as an example because after removing only $15$ (approximately $1/10,000$) high-score features, it is corrected into its true class, kite, suggesting the high contribution of the $15$ high-score features to the reason for misclassification. We transform the removed Fourier features back to the spatial domain and observe that the deleted signals, which cause the misclassification, exhibit visual similarities to class bustard (the original misclassified class). In this example, both the deleted features and the samples of bustard display a yellow background. In contrast, most samples in class kite have a blue sky background, while class bustard contains more samples with yellow backgrounds (Figure~\ref{fig:experiment}.L). Based on this, we infer that the network misclassifies the sample as it incorrectly extracts the feature of the yellow background and associates it with the class bustard, which genuinely has yellow background features. Which means that the network might mistakenly take the background information as a feature of the bustard. Using  FFC, we successfully and interpretably correct the network's decision. This example further demonstrates that even though FFC conducts attribution in the Fourier domain, the corresponding spatial features can still be visually perceived by human beings.
\section{Conclusion}
This paper demonstrates the rationale for applying game theory-based metrics in the context of Fourier feature attribution. Then it presents the Fast Fourier Correlation (FFC) attribution algorithm method based on control theory, providing a novel perspective for feature attribution in the Fourier Domain. Experimental results validate the effectiveness of the proposed method, showcasing the superior feature selection capabilities, as well as the difference from the simple FFT/IFFT of any existing works. By comparing spatial domain attribution results with those of Fourier domain attribution, experiments show that Fourier features exhibit better intra-class concentration and comparable inter-class specificity, indicating that Fourier features are more suitable for classification tasks and the development of explainable AI. Finally, the paper visualizes the Fourier features via misclassification correction analysis, showing that the corresponding spatial features can be visually perceived by human beings.
Although the paper demonstrates the possibility of leveraging FFC to interpretably correct misclassification, it didn't systematically evaluate the relationship between the Fourier features found by FFC and the reason for misclassification. Thus, further studies might provide insight into this field based on FFC and shed light on interpretable AI. Meanwhile, related works might contribute greatly to the field of backdoor detection or adversary sample generation.
Based on FFC, more studies could aim to develop explainable AI, enhancing the reliability and robustness of AI and complementing the shortcomings in the current attribution methods.
\bibliography{ref}

\begin{thebibliography}{52}
\providecommand{\natexlab}[1]{#1}
\providecommand{\url}[1]{\texttt{#1}}
\expandafter\ifx\csname urlstyle\endcsname\relax
  \providecommand{\doi}[1]{doi: #1}\else
  \providecommand{\doi}{doi: \begingroup \urlstyle{rm}\Url}\fi

\bibitem[Adebayo et~al.(2020)Adebayo, Muelly, Liccardi, and Kim]{unstablenips2020}
Adebayo, J., Muelly, M., Liccardi, I., and Kim, B.
\newblock Debugging tests for model explanations.
\newblock In Larochelle, H., Ranzato, M., Hadsell, R., Balcan, M., and Lin, H. (eds.), \emph{Advances in Neural Information Processing Systems 33: Annual Conference on Neural Information Processing Systems 2020, NeurIPS 2020, December 6-12, 2020, virtual}, 2020.

\bibitem[Amara et~al.(2022)Amara, Ying, Zhang, Han, Zhao, Shan, Brandes, Schemm, and Zhang]{FID}
Amara, K., Ying, Z., Zhang, Z., Han, Z., Zhao, Y., Shan, Y., Brandes, U., Schemm, S., and Zhang, C.
\newblock Graphframex: Towards systematic evaluation of explainability methods for graph neural networks.
\newblock In Rieck, B. and Pascanu, R. (eds.), \emph{Learning on Graphs Conference, LoG 2022, 9-12 December 2022, Virtual Event}, volume 198 of \emph{Proceedings of Machine Learning Research}, pp.\ ~44. {PMLR}, 2022.

\bibitem[Ancona et~al.(2017)Ancona, Ceolini, {\"{O}}ztireli, and Gross]{metricassumption}
Ancona, M., Ceolini, E., {\"{O}}ztireli, A.~C., and Gross, M.~H.
\newblock A unified view of gradient-based attribution methods for deep neural networks.
\newblock \emph{CoRR}, abs/1711.06104, 2017.

\bibitem[Bilodeau et~al.(2022)Bilodeau, Jaques, Koh, and Kim]{unstabletheo22}
Bilodeau, B.~L., Jaques, N., Koh, P.~W., and Kim, B.
\newblock Impossibility theorems for feature attribution.
\newblock \emph{CoRR}, abs/2212.11870, 2022.
\newblock \doi{10.48550/ARXIV.2212.11870}.
\newblock URL \url{https://doi.org/10.48550/arXiv.2212.11870}.

\bibitem[Chen et~al.(2022)Chen, Ren, and Yan]{shapFre}
Chen, Y., Ren, Q., and Yan, J.
\newblock Rethinking and improving robustness of convolutional neural networks: a shapley value-based approach in frequency domain.
\newblock In Koyejo, S., Mohamed, S., Agarwal, A., Belgrave, D., Cho, K., and Oh, A. (eds.), \emph{Advances in Neural Information Processing Systems}, pp.\  324--337. Curran Associates, Inc., 2022.

\bibitem[Dabkowski \& Gal(2017)Dabkowski and Gal]{metricassumtionnips}
Dabkowski, P. and Gal, Y.
\newblock Real time image saliency for black box classifiers.
\newblock In Guyon, I., von Luxburg, U., Bengio, S., Wallach, H.~M., Fergus, R., Vishwanathan, S. V.~N., and Garnett, R. (eds.), \emph{Advances in Neural Information Processing Systems 30: Annual Conference on Neural Information Processing Systems 2017, December 4-9, 2017, Long Beach, CA, {USA}}, pp.\  6967--6976, 2017.

\bibitem[Decker et~al.(2024)Decker, Bhattarai, Gu, Tresp, and Buettner]{aggattri}
Decker, T., Bhattarai, A.~R., Gu, J., Tresp, V., and Buettner, F.
\newblock Provably better explanations with optimized aggregation of feature attributions.
\newblock In \emph{Forty-first International Conference on Machine Learning, {ICML} 2024, Vienna, Austria, July 21-27, 2024}. OpenReview.net, 2024.

\bibitem[Deng et~al.(2024)Deng, Zou, Du, Chen, Feng, Yang, Li, and Zhang]{Quanshi}
Deng, H., Zou, N., Du, M., Chen, W., Feng, G., Yang, Z., Li, Z., and Zhang, Q.
\newblock Unifying fourteen post-hoc attribution methods with taylor interactions.
\newblock \emph{IEEE Transactions on Pattern Analysis and Machine Intelligence}, 46\penalty0 (7), 2024.
\newblock \doi{10.1109/TPAMI.2024.3358410}.

\bibitem[Dombrowski et~al.(2019)Dombrowski, Alber, Anders, Ackermann, M{\"{u}}ller, and Kessel]{unstable}
Dombrowski, A., Alber, M., Anders, C.~J., Ackermann, M., M{\"{u}}ller, K., and Kessel, P.
\newblock Explanations can be manipulated and geometry is to blame.
\newblock In Wallach, H.~M., Larochelle, H., Beygelzimer, A., d'Alch{\'{e}}{-}Buc, F., Fox, E.~B., and Garnett, R. (eds.), \emph{Advances in Neural Information Processing Systems 32: Annual Conference on Neural Information Processing Systems 2019, NeurIPS 2019, December 8-14, 2019, Vancouver, BC, Canada}, pp.\  13567--13578, 2019.

\bibitem[Dong et~al.(2021)Dong, Cordonnier, and Loukas]{vitlowfre}
Dong, Y., Cordonnier, J., and Loukas, A.
\newblock Attention is not all you need: pure attention loses rank doubly exponentially with depth.
\newblock In Meila, M. and Zhang, T. (eds.), \emph{Proceedings of the 38th International Conference on Machine Learning, {ICML} 2021, 18-24 July 2021, Virtual Event}, volume 139 of \emph{Proceedings of Machine Learning Research}, pp.\  2793--2803. {PMLR}, 2021.

\bibitem[Erion et~al.(2021)Erion, Janizek, Sturmfels, Lundberg, and Lee]{EG}
Erion, G.~G., Janizek, J.~D., Sturmfels, P., Lundberg, S.~M., and Lee, S.
\newblock Improving performance of deep learning models with axiomatic attribution priors and expected gradients.
\newblock \emph{Nat. Mach. Intell.}, 3\penalty0 (7), 2021.
\newblock \doi{10.1038/S42256-021-00343-W}.

\bibitem[Fel et~al.(2023)Fel, Picard, B{\'{e}}thune, Boissin, Vigouroux, Colin, Cad{\`{e}}ne, and Serre]{craftartifact}
Fel, T., Picard, A.~M., B{\'{e}}thune, L., Boissin, T., Vigouroux, D., Colin, J., Cad{\`{e}}ne, R., and Serre, T.
\newblock {CRAFT:} concept recursive activation factorization for explainability.
\newblock In \emph{{IEEE/CVF} Conference on Computer Vision and Pattern Recognition, {CVPR} 2023}, 2023.
\newblock \doi{10.1109/CVPR52729.2023.00266}.

\bibitem[Fokkema et~al.(2023)Fokkema, de~Heide, and van Erven]{unstabletheo23}
Fokkema, H., de~Heide, R., and van Erven, T.
\newblock Attribution-based explanations that provide recourse cannot be robust.
\newblock \emph{J. Mach. Learn. Res.}, 24:\penalty0 360:1--360:37, 2023.

\bibitem[Gerstner et~al.(2014)Gerstner, Kistler, Naud, and Paninski]{neurondynamics}
Gerstner, W., Kistler, W.~M., Naud, R., and Paninski, L.
\newblock \emph{Neuronal dynamics: From single neurons to networks and models of cognition}.
\newblock Cambridge University Press, 2014.

\bibitem[Hooker et~al.(2019)Hooker, Erhan, Kindermans, and Kim]{attbenchmark}
Hooker, S., Erhan, D., Kindermans, P., and Kim, B.
\newblock A benchmark for interpretability methods in deep neural networks.
\newblock In Wallach, H.~M., Larochelle, H., Beygelzimer, A., d'Alch{\'{e}}{-}Buc, F., Fox, E.~B., and Garnett, R. (eds.), \emph{Advances in Neural Information Processing Systems 32: Annual Conference on Neural Information Processing Systems 2019, NeurIPS 2019, December 8-14, 2019, Vancouver, BC, Canada}, pp.\  9734--9745, 2019.

\bibitem[Jianxun~Zhao(2018)]{ampli-filter}
Jianxun~Zhao, J.~D.
\newblock \emph{Fundamentals of Radio Frequency Circuits}.
\newblock Xidian University Press, 2018.

\bibitem[Kim et~al.(2018)Kim, Wattenberg, Gilmer, Cai, Wexler, Vi{\'{e}}gas, and Sayres]{TCAV}
Kim, B., Wattenberg, M., Gilmer, J., Cai, C.~J., Wexler, J., Vi{\'{e}}gas, F.~B., and Sayres, R.
\newblock Interpretability beyond feature attribution: Quantitative testing with concept activation vectors {(TCAV)}.
\newblock In Dy, J.~G. and Krause, A. (eds.), \emph{Proceedings of the 35th International Conference on Machine Learning, {ICML} 2018}, volume~80 of \emph{Proceedings of Machine Learning Research}, 2018.

\bibitem[Kokhlikyan et~al.(2020)Kokhlikyan, Miglani, Martin, Wang, Alsallakh, Reynolds, Melnikov, Kliushkina, Araya, Yan, and Reblitz{-}Richardson]{captum}
Kokhlikyan, N., Miglani, V., Martin, M., Wang, E., Alsallakh, B., Reynolds, J., Melnikov, A., Kliushkina, N., Araya, C., Yan, S., and Reblitz{-}Richardson, O.
\newblock Captum: {A} unified and generic model interpretability library for pytorch.
\newblock \emph{CoRR}, abs/2009.07896, 2020.

\bibitem[Krishna et~al.(2024)Krishna, Han, Gu, Wu, Jabbari, and Lakkaraju]{conflict1}
Krishna, S., Han, T., Gu, A., Wu, S., Jabbari, S., and Lakkaraju, H.
\newblock The disagreement problem in explainable machine learning: {A} practitioner's perspective.
\newblock \emph{Trans. Mach. Learn. Res.}, 2024, 2024.

\bibitem[Ley et~al.(2023)Ley, Mishra, and Magazzeni]{harvardTrajectory}
Ley, D., Mishra, S., and Magazzeni, D.
\newblock {GLOBE-CE:} {A} translation based approach for global counterfactual explanations.
\newblock In Krause, A., Brunskill, E., Cho, K., Engelhardt, B., Sabato, S., and Scarlett, J. (eds.), \emph{International Conference on Machine Learning, {ICML} 2023}, volume 202 of \emph{Proceedings of Machine Learning Research}, 2023.
\newblock URL \url{https://proceedings.mlr.press/v202/ley23a.html}.

\bibitem[Lin et~al.(2023)Lin, Covert, and Lee]{unstablenips}
Lin, C., Covert, I., and Lee, S.
\newblock On the robustness of removal-based feature attributions.
\newblock In Oh, A., Naumann, T., Globerson, A., Saenko, K., Hardt, M., and Levine, S. (eds.), \emph{Advances in Neural Information Processing Systems 36: Annual Conference on Neural Information Processing Systems 2023, NeurIPS 2023, New Orleans, LA, USA, December 10 - 16, 2023}, 2023.

\bibitem[Ma et~al.(2021)Ma, Xu, and Zhang]{Fprinciple}
Ma, Y., Xu, Z.~J., and Zhang, J.
\newblock Frequency principle in deep learning beyond gradient-descent-based training.
\newblock \emph{CoRR}, abs/2101.00747, 2021.

\bibitem[Marcellin et~al.(2000)Marcellin, Gormish, Bilgin, and Boliek]{compresstheory}
Marcellin, M., Gormish, M., Bilgin, A., and Boliek, M.
\newblock An overview of jpeg-2000.
\newblock In \emph{Proceedings DCC 2000. Data Compression Conference}, pp.\  523--541, 2000.
\newblock \doi{10.1109/DCC.2000.838192}.

\bibitem[Mi et~al.(2023)Mi, Le, He, Shlizerman, and S{\"u}mb{\"u}l]{analyneurondynamics1}
Mi, L., Le, T., He, T., Shlizerman, E., and S{\"u}mb{\"u}l, U.
\newblock Learning time-invariant representations for individual neurons from population dynamics.
\newblock In \emph{Thirty-seventh Conference on Neural Information Processing Systems}, 2023.

\bibitem[Neely et~al.(2021)Neely, Schouten, Bleeker, and Lucic]{conflict2}
Neely, M., Schouten, S.~F., Bleeker, M. J.~R., and Lucic, A.
\newblock Order in the court: Explainable {AI} methods prone to disagreement.
\newblock \emph{CoRR}, abs/2105.03287, 2021.

\bibitem[Parmar et~al.(2019)Parmar, Ramachandran, Vaswani, Bello, Levskaya, and Shlens]{vitgood}
Parmar, N., Ramachandran, P., Vaswani, A., Bello, I., Levskaya, A., and Shlens, J.
\newblock Stand-alone self-attention in vision models.
\newblock In Wallach, H.~M., Larochelle, H., Beygelzimer, A., d'Alch{\'{e}}{-}Buc, F., Fox, E.~B., and Garnett, R. (eds.), \emph{Advances in Neural Information Processing Systems 32: Annual Conference on Neural Information Processing Systems 2019, NeurIPS 2019, December 8-14, 2019, Vancouver, BC, Canada}, pp.\  68--80, 2019.

\bibitem[Rahaman et~al.(2019)Rahaman, Baratin, Arpit, Draxler, Lin, Hamprecht, Bengio, and Courville]{icmlciteFpreciple}
Rahaman, N., Baratin, A., Arpit, D., Draxler, F., Lin, M., Hamprecht, F.~A., Bengio, Y., and Courville, A.~C.
\newblock On the spectral bias of neural networks.
\newblock In Chaudhuri, K. and Salakhutdinov, R. (eds.), \emph{Proceedings of the 36th International Conference on Machine Learning, {ICML} 2019, 9-15 June 2019, Long Beach, California, {USA}}, volume~97 of \emph{Proceedings of Machine Learning Research}, pp.\  5301--5310. {PMLR}, 2019.

\bibitem[Ribeiro et~al.(2016)Ribeiro, Singh, and Guestrin]{LIME}
Ribeiro, M.~T., Singh, S., and Guestrin, C.
\newblock "why should {I} trust you?": Explaining the predictions of any classifier.
\newblock In Krishnapuram, B., Shah, M., Smola, A.~J., Aggarwal, C.~C., Shen, D., and Rastogi, R. (eds.), \emph{Proceedings of the 22nd {ACM} {SIGKDD} International Conference on Knowledge Discovery and Data Mining, San Francisco, CA, USA, August 13-17, 2016}, pp.\  1135--1144. {ACM}, 2016.
\newblock \doi{10.1145/2939672.2939778}.

\bibitem[Rieger \& Hansen(2020)Rieger and Hansen]{IR}
Rieger, L. and Hansen, L.~K.
\newblock {IROF:} a low resource evaluation metric for explanation methods.
\newblock \emph{CoRR}, abs/2003.08747, 2020.

\bibitem[Russakovsky et~al.(2015)Russakovsky, Deng, Su, Krause, Satheesh, Ma, Huang, Karpathy, Khosla, Bernstein, Berg, and Fei{-}Fei]{imgnet2012}
Russakovsky, O., Deng, J., Su, H., Krause, J., Satheesh, S., Ma, S., Huang, Z., Karpathy, A., Khosla, A., Bernstein, M.~S., Berg, A.~C., and Fei{-}Fei, L.
\newblock Imagenet large scale visual recognition challenge.
\newblock \emph{Int. J. Comput. Vis.}, 115\penalty0 (3), 2015.
\newblock \doi{10.1007/S11263-015-0816-Y}.

\bibitem[Selvaraju et~al.(2017)Selvaraju, Cogswell, Das, Vedantam, Parikh, and Batra]{gradcam}
Selvaraju, R.~R., Cogswell, M., Das, A., Vedantam, R., Parikh, D., and Batra, D.
\newblock Grad-cam: Visual explanations from deep networks via gradient-based localization.
\newblock In \emph{{IEEE} International Conference on Computer Vision, {ICCV} 2017}, 2017.
\newblock \doi{10.1109/ICCV.2017.74}.

\bibitem[Shrikumar et~al.(2017)Shrikumar, Greenside, and Kundaje]{deeplift}
Shrikumar, A., Greenside, P., and Kundaje, A.
\newblock Learning important features through propagating activation differences.
\newblock In Precup, D. and Teh, Y.~W. (eds.), \emph{Proceedings of the 34th International Conference on Machine Learning, {ICML} 2017}, volume~70 of \emph{Proceedings of Machine Learning Research}, 2017.

\bibitem[Simonyan et~al.(2014)Simonyan, Vedaldi, and Zisserman]{inputgrad}
Simonyan, K., Vedaldi, A., and Zisserman, A.
\newblock Deep inside convolutional networks: Visualising image classification models and saliency maps.
\newblock In Bengio, Y. and LeCun, Y. (eds.), \emph{2nd International Conference on Learning Representations, {ICLR} 2014}, 2014.

\bibitem[Smilkov et~al.(2017)Smilkov, Thorat, Kim, Vi{\'{e}}gas, and Wattenberg]{smoothgrad}
Smilkov, D., Thorat, N., Kim, B., Vi{\'{e}}gas, F.~B., and Wattenberg, M.
\newblock Smoothgrad: removing noise by adding noise.
\newblock \emph{CoRR}, abs/1706.03825, 2017.
\newblock URL \url{http://arxiv.org/abs/1706.03825}.

\bibitem[Smith(1997)]{Conv}
Smith, S.~W.
\newblock \emph{The Scientist and Engineer Guide to Digital Signal Processing}.
\newblock California Technical Publishing, 1997.

\bibitem[Springenberg et~al.(2015)Springenberg, Dosovitskiy, Brox, and Riedmiller]{gbp}
Springenberg, J.~T., Dosovitskiy, A., Brox, T., and Riedmiller, M.~A.
\newblock Striving for simplicity: The all convolutional net.
\newblock In Bengio, Y. and LeCun, Y. (eds.), \emph{3rd International Conference on Learning Representations, {ICLR} 2015, San Diego, CA, USA, May 7-9, 2015, Workshop Track Proceedings}, 2015.

\bibitem[Srinivas \& Fleuret(2019)Srinivas and Fleuret]{fullgrad}
Srinivas, S. and Fleuret, F.
\newblock Full-gradient representation for neural network visualization.
\newblock In Wallach, H.~M., Larochelle, H., Beygelzimer, A., d'Alch{\'{e}}{-}Buc, F., Fox, E.~B., and Garnett, R. (eds.), \emph{Advances in Neural Information Processing Systems 32: Annual Conference on Neural Information Processing Systems 2019, NeurIPS 2019}, 2019.

\bibitem[Sundararajan et~al.(2017)Sundararajan, Taly, and Yan]{intgrad}
Sundararajan, M., Taly, A., and Yan, Q.
\newblock Axiomatic attribution for deep networks.
\newblock In Precup, D. and Teh, Y.~W. (eds.), \emph{Proceedings of the 34th International Conference on Machine Learning, {ICML} 2017}, volume~70 of \emph{Proceedings of Machine Learning Research}, 2017.

\bibitem[Tancik et~al.(2020)Tancik, Srinivasan, Mildenhall, Fridovich-Keil, Raghavan, Singhal, Ramamoorthi, Barron, and Ng]{NTKFourier}
Tancik, M., Srinivasan, P.~P., Mildenhall, B., Fridovich-Keil, S., Raghavan, N., Singhal, U., Ramamoorthi, R., Barron, J.~T., and Ng, R.
\newblock Fourier features let networks learn high frequency functions in low dimensional domains.
\newblock In \emph{Proceedings of the 34th International Conference on Neural Information Processing Systems}. Curran Associates Inc., 2020.

\bibitem[Wang et~al.(2022{\natexlab{a}})Wang, Zheng, Chen, and Wang]{wang2022antioversmoothing}
Wang, P., Zheng, W., Chen, T., and Wang, Z.
\newblock Anti-oversmoothing in deep vision transformers via the fourier domain analysis: From theory to practice.
\newblock In \emph{International Conference on Learning Representations}, 2022{\natexlab{a}}.

\bibitem[Wang et~al.(2022{\natexlab{b}})Wang, Luo, WANG, Ding, Wang, and Li]{modelcomression}
Wang, Z., Luo, H., WANG, P., Ding, F., Wang, F., and Li, H.
\newblock Vtc-lfc: Vision transformer compression with low-frequency components.
\newblock In Koyejo, S., Mohamed, S., Agarwal, A., Belgrave, D., Cho, K., and Oh, A. (eds.), \emph{Advances in Neural Information Processing Systems}, pp.\  13974--13988, 2022{\natexlab{b}}.

\bibitem[Wen \& Jacot(2024)Wen and Jacot]{wen2024which}
Wen, Y. and Jacot, A.
\newblock Which frequencies do {CNN}s need? emergent bottleneck structure in feature learning.
\newblock In \emph{Forty-first International Conference on Machine Learning}, 2024.

\bibitem[Xu \& Zhou(2021)Xu and Zhou]{zhiqinxuintroDFprinciple}
Xu, Z.~J. and Zhou, H.
\newblock Deep frequency principle towards understanding why deeper learning is faster.
\newblock In \emph{Thirty-Fifth {AAAI} Conference on Artificial Intelligence, {AAAI} 2021, Thirty-Third Conference on Innovative Applications of Artificial Intelligence, {IAAI} 2021, The Eleventh Symposium on Educational Advances in Artificial Intelligence, {EAAI} 2021, Virtual Event, February 2-9, 2021}, pp.\  10541--10550. {AAAI} Press, 2021.
\newblock \doi{10.1609/AAAI.V35I12.17261}.

\bibitem[Xu et~al.(2019)Xu, Zhang, and Xiao]{frequencydomaintrainingbehavior}
Xu, Z.~J., Zhang, Y., and Xiao, Y.
\newblock Training behavior of deep neural network in frequency domain.
\newblock In Gedeon, T., Wong, K.~W., and Lee, M. (eds.), \emph{Neural Information Processing - 26th International Conference, {ICONIP} 2019}, 2019.
\newblock \doi{10.1007/978-3-030-36708-4\_22}.

\bibitem[Yang et~al.(2023{\natexlab{a}})Yang, Akhtar, Wen, and Mian]{LPI}
Yang, P., Akhtar, N., Wen, Z., and Mian, A.
\newblock Local path integration for attribution.
\newblock In Williams, B., Chen, Y., and Neville, J. (eds.), \emph{Thirty-Seventh {AAAI} Conference on Artificial Intelligence, {AAAI} 2023, Thirty-Fifth Conference on Innovative Applications of Artificial Intelligence}, 2023{\natexlab{a}}.
\newblock \doi{10.1609/AAAI.V37I3.25422}.

\bibitem[Yang et~al.(2023{\natexlab{b}})Yang, Akhtar, Wen, Shah, and Mian]{DiffID2}
Yang, P., Akhtar, N., Wen, Z., Shah, M., and Mian, A.~S.
\newblock Re-calibrating feature attributions for model interpretation.
\newblock In \emph{The Eleventh International Conference on Learning Representations, {ICLR} 2023}, 2023{\natexlab{b}}.

\bibitem[Yeh et~al.(2019)Yeh, Hsieh, Suggala, Inouye, and Ravikumar]{infd}
Yeh, C., Hsieh, C., Suggala, A.~S., Inouye, D.~I., and Ravikumar, P.
\newblock On the (in)fidelity and sensitivity of explanations.
\newblock In Wallach, H.~M., Larochelle, H., Beygelzimer, A., d'Alch{\'{e}}{-}Buc, F., Fox, E.~B., and Garnett, R. (eds.), \emph{Advances in Neural Information Processing Systems 32: Annual Conference on Neural Information Processing Systems 2019, NeurIPS 2019, December 8-14, 2019, Vancouver, BC, Canada}, pp.\  10965--10976, 2019.

\bibitem[Yin et~al.(2019)Yin, Lopes, Shlens, Cubuk, and Gilmer]{frequency1robustness}
Yin, D., Lopes, R.~G., Shlens, J., Cubuk, E.~D., and Gilmer, J.
\newblock A fourier perspective on model robustness in computer vision.
\newblock In Wallach, H.~M., Larochelle, H., Beygelzimer, A., d'Alch{\'{e}}{-}Buc, F., Fox, E.~B., and Garnett, R. (eds.), \emph{Advances in Neural Information Processing Systems 32: Annual Conference on Neural Information Processing Systems 2019, NeurIPS 2019, December 8-14, 2019, Vancouver, BC, Canada}, pp.\  13255--13265, 2019.

\bibitem[Zaher et~al.(2024)Zaher, Trzaskowski, Nguyen, and Roosta]{MIG}
Zaher, E., Trzaskowski, M., Nguyen, Q., and Roosta, F.
\newblock Manifold integrated gradients: Riemannian geometry for feature attribution.
\newblock In \emph{Forty-first International Conference on Machine Learning, {ICML} 2024, Vienna, Austria, July 21-27, 2024}. OpenReview.net, 2024.

\bibitem[Zhou et~al.(2018)Zhou, Sun, Bau, and Torralba]{FromTCAV}
Zhou, B., Sun, Y., Bau, D., and Torralba, A.
\newblock Interpretable basis decomposition for visual explanation.
\newblock In Ferrari, V., Hebert, M., Sminchisescu, C., and Weiss, Y. (eds.), \emph{Computer Vision - {ECCV} 2018 - 15th European Conference}, volume 11212 of \emph{Lecture Notes in Computer Science}, 2018.
\newblock \doi{10.1007/978-3-030-01237-3\_8}.

\bibitem[Zhou et~al.(2022)Zhou, Booth, Ribeiro, and Shah]{unstableAAAI}
Zhou, Y., Booth, S., Ribeiro, M.~T., and Shah, J.
\newblock Do feature attribution methods correctly attribute features?
\newblock In \emph{Thirty-Sixth {AAAI} Conference on Artificial Intelligence, {AAAI} 2022, Thirty-Fourth Conference on Innovative Applications of Artificial Intelligence, {IAAI} 2022, The Twelveth Symposium on Educational Advances in Artificial Intelligence, {EAAI} 2022 Virtual Event, February 22 - March 1, 2022}, pp.\  9623--9633. {AAAI} Press, 2022.
\newblock \doi{10.1609/AAAI.V36I9.21196}.

\bibitem[Zintgraf et~al.(2017)Zintgraf, Cohen, Adel, and Welling]{withBenchmark1}
Zintgraf, L.~M., Cohen, T.~S., Adel, T., and Welling, M.
\newblock Visualizing deep neural network decisions: Prediction difference analysis.
\newblock In \emph{5th International Conference on Learning Representations, {ICLR} 2017, Toulon, France, April 24-26, 2017, Conference Track Proceedings}. OpenReview.net, 2017.

\end{thebibliography}
\bibliographystyle{refer}

\end{document}